\documentclass{article}

\usepackage{arxiv}
\usepackage[utf8]{inputenc} % allow utf-8 input
\usepackage[T1]{fontenc}    % use 8-bit T1 fonts
\usepackage{hyperref}
\usepackage{url}            % simple URL typesetting
\usepackage{booktabs}       % professional-quality tables
\usepackage{amsfonts}       % blackboard math symbols
\usepackage{nicefrac}       % compact symbols for 1/2, etc.
\usepackage[numbers]{natbib}
\usepackage{microtype}      % microtypography
\usepackage{graphicx}
\usepackage{amsmath,amsthm,amsfonts,amssymb}
\usepackage{caption}
\usepackage{float}
\usepackage{subfig}
\usepackage{bm}
\title{Integrating Fourier Neural Operators with Diffusion Models to improve Spectral Representation of Synthetic Earthquake Ground Motion Response}

\author{
    Niccolò Perrone\\ 
    Politecnico di Milano\\
    P.zza Leonardo da Vinci 32, I-20133, Milano, Italy,\\
    \texttt{niccolo.perrone@mail.polimi.it} \\
    \And 
    Fanny Lehmann\\
    ETH AI Center \& Seminar for Applied Mathematics\\
    Zurich, Switzerland \\
    \texttt{fanny.lehmann@ai.ethz.ch}\\
    \And
    Hugo Gabrielidis\\
    Universit\'e Paris-Saclay\\
    CentraleSup\'elec, CNRS, ENS Paris-Saclay\\
    Laboratoire de Mécanique Paris-Saclay UMR 9026\\
    8-10 rue Joliot Curie, 91190 Gif-sur-Yvette, France\\
    \texttt{hugo.gabrielidis@centralesupelec.fr}\\
    \And
    Stefania Fresca\\
    MOX - Department of Mathematics\\
    Politecnico di Milano\\
    P.zza Leonardo da Vinci 32, I-20133, Milano, Italy\\
    \texttt{stefania.fresca@polimi.it}\\
    \And
    Filippo Gatti
    Universit\'e Paris-Saclay\\
    CentraleSup\'elec, CNRS, ENS Paris-Saclay\\
    Laboratoire de Mécanique Paris-Saclay UMR 9026\\
    8-10 rue Joliot Curie, 91190 Gif-sur-Yvette, France\\
    \texttt{filippo.gatti@centralesupelec.fr}
}
\date{\today}

\newcommand{\SEM}{\href{https://github.com/sem3d/sem}{SEM3D}}
\newcommand{\MIFNO}{\href{https://github.com/lehmannfa/MIFNO}{MIFNO}}
\newcommand{\DDPM}{DDPM}
\newcommand{\MIFNODDPM}{MIFNO+DDPM}
\newcommand{\HEMEWS}{\href{https://entrepot.recherche.data.gouv.fr/dataset.xhtml?persistentId=doi:10.57745/LAI6YU}{HEMEW$^\text{{S}}${-3D}}}

\begin{document}

\maketitle

% \vspace{-0.8cm}
% \begin{center}
%     Submitted at the 28th International Conference on Structural Mechanics in Reactor Technology\\
%     Toronto, Canada, August 10-15, 2025\\
%     Division XI
% \end{center}

\begin{abstract}
Nuclear reactor buildings must be designed to withstand the dynamic load induced by strong ground motion earthquakes. For this reason, their structural behavior must be assessed in multiple realistic ground shaking scenarios (\textit{e.g.}, the Maximum Credible Earthquake). However, earthquake catalogs and recorded seismograms may not always be available in the region of interest. Therefore, synthetic earthquake ground motion is progressively being employed, although with some due precautions: earthquake physics is sometimes not well enough understood to be accurately reproduced with numerical tools, and the underlying epistemic uncertainties lead to prohibitive computational costs related to model calibration. In this study, we propose an AI physics-based approach to generate synthetic ground motion, based on the combination of a neural operator that approximates the elastodynamics Green's operator in arbitrary source-geology setups, enhanced by a denoising diffusion probabilistic model. The diffusion model is trained to correct the ground motion time series generated by the neural operator.
Our results show that such an approach promisingly enhances the realism of the generated synthetic seismograms, with frequency biases and Goodness-Of-Fit (GOF) scores being improved by the diffusion model. This indicates that the latter is capable to mitigate the mid-frequency spectral falloff observed in the time series generated by the neural operator. Our method showcases fast and cheap inference in different site and source conditions.
\end{abstract}


\section{Introduction}
\label{sec-introduction}
Recent advances in high-performance computing (HPC) tools have enabled the numerical simulation of elastodynamic problems in complex setups. In this context, simulating large-scale earthquake scenarios, typically spanning urban regions ($\approx$100 km $\times$ 100 km $\times$ 50 km), requires high-resolution models with sub-meter accuracy, to match the Earth's complex structure. Current state-of-the-art simulations can model active fault ruptures and wave propagation through complex geological structures, but they still face challenges in simulating high-frequency surface motions, limiting the fidelity of fault-to-structure and soil-structure interaction analyses~\citep{Fares_CastroCruz_Foerster_LopezCaballero_Gatti_2022}. These issues arise not only due to the inherent computational demand, but especially because of the uncertainties in subsurface properties~\citep{CastroCruz_Gatti_LopezCaballero_2021a} and seismogenetic source~\citep{Nakao_Ichimura_Fujita_Hori_Kobayashi_Munekane_2024}, making it difficult to generate broadband signals suitable for structural design.

The latter task of generating broadband (0-30 Hz) seismograms, is still far from being accomplished, either with empirical~\citep{Kuehn_Campbell_Bozorgnia_2025}, or numerical approaches~\citep{Graves_Pitarka_2010}. With the recent machine learning improvements, neural networks have been successfully applied to enhancing low-frequency seismograms with high-frequency content~\citep{florezDataDrivenSynthesisBroadband2022}. Among the developed methods, diffusion models~\citep{Ho_et_al_2020} stand out due to their ease of training, performance, diversity of inferred samples, and the ability to use a variety of preexisting architectures as backbone generator. Diffusion models are increasingly being applied to generate realistic ground motion scenarios, conditioned on various metadata such as site and event characteristics (\textit{e.g.}, magnitude, source-to-site distance, and coordinates, \cite{Bergmeister_Palgunadi_Bosisio_Ermert_Koroni_Perraudin_Dirmeier_Meier_2024}, among others).\\

In this study, we present a novel deep learning strategy to yield multiple realistic strong-motion accelerograms in the 0-5 Hz frequency range, combining a physics-based surrogate model with a generative model. To this end, we employ the Multiple-Input Fourier Neural Operator (MIFNO) proposed by~\cite{Lehmann_Gatti_Clouteau_2025} in conjunction with the conditional diffusion architecture presented in~\cite{Gabrielidis_Gatti_Vialle_Jacquet_2024}. The \MIFNO\ acts as a fast surrogate model of 3D elastodynamics for arbitrary geological settings and source characteristics. We address \MIFNO\ limitations in capturing mid-frequency details by employing a denoising diffusion probabilistic model~\citep[DDPM,][]{Ho_et_al_2020} to improve the spectral content of the inferred wavefield at the free surface. Both the \MIFNO\ and the \DDPM\ are trained on elastodynamics simulations, the targets being the time-dependent wavefields for the \MIFNO\ and the single-station seismograms for the \DDPM. %, the \DDPM\ is trained on thousands of seismograms recorded worldwide, so to learn the single-station low-to-high frequency mapping and improve the realism of the \MIFNO\ output at inference time. 
The results show that the combination of \MIFNO\ and \DDPM\ increases the realism of the rendered elastodynamic wavefield, by mitigating the \MIFNO\ mid-frequency spectral bias. Our method preserves the fast inference of multiple earthquake scenarios (any geology, any source).
The remainder of the paper is organized as follows: Section METHODS describes the key concepts behind \MIFNO\ and \DDPM. The results are discussed in Section RESULTS, and the conclusions are drawn in Section CONCLUSIONS.

\section{Methods}
\label{sec-methods}
\subsection{Multiple-Input Fourier Neural Operator (MIFNO)}
\label{subsec-mifno}
The 3D elastodynamics problem in a domain $\Omega\subset\mathbb{R}^3$ reads:
\begin{equation}
    \mathcal{L}\left[\bm{a},\bm{v}\right](\bm{x},t) = \bm{f}[\bm{x_s}, \bm{\theta_s}](\bm{x}, t),\quad \forall (\bm{x},t)\in\Omega\times\left]0,T\right]
    \label{eq:general_equation}
\end{equation}
where $\mathcal{L}\left[\bm{a},\bm{v}\right]$ is the 3D elastic wave equation parametrized by the Lamé parameters $\bm{a}(\bm{x})$. $\bm{f} \in L^2(\Omega \times ]0, T[,\mathbb{R}^3)$ is the body force distribution, expressed as $\mathrm{div}~\:~\bm{m}(\bm{x})~\cdot~s(t)$, the divergence of a double-couple moment tensor density $\bm{m}\in H^1\left(\Omega\right)^{6}$, localized at a point-wise location $\bm{x_s}$ and with strike, dip, rake angles regrouped in vector $\bm{\theta_s}\in \mathbb{R}^3$. $s(t)\in C^0\left(\left]0,T\right]\right)$ represents the time-varying source amplitude over the duration $T$. Finally, $\bm{v}\in \mathcal{C}^1(\Omega \times ]0, T[,\mathbb{R}^3)$ is the particle velocity field, solution of the problem.

\begin{figure}[ht!]
    \centering
    \includegraphics[width=\textwidth]{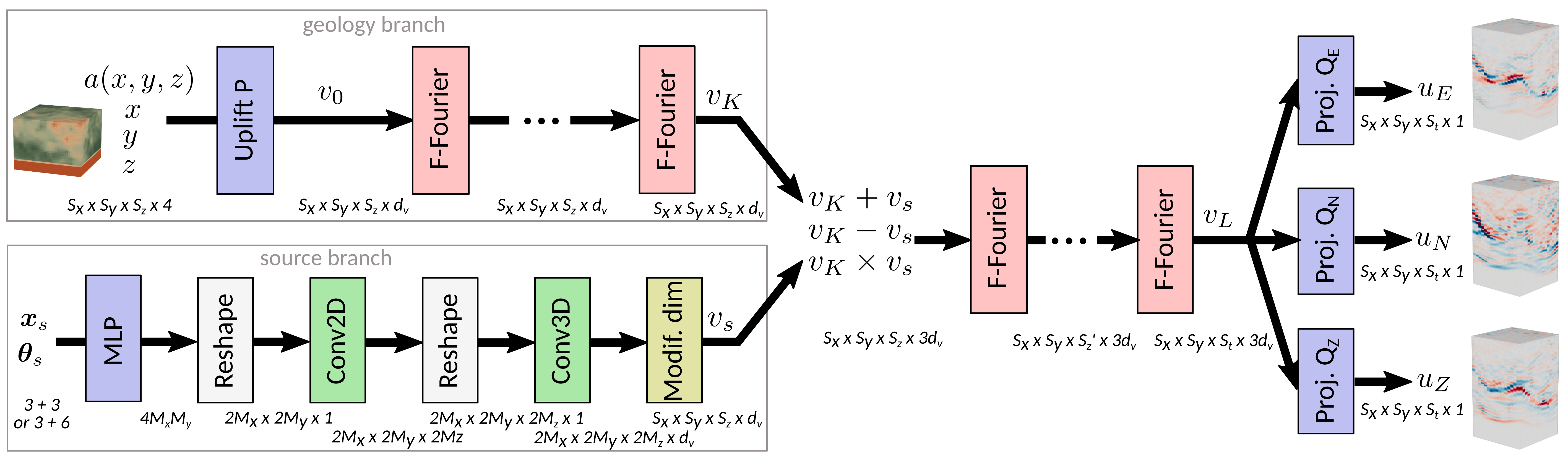}
    \caption{The \MIFNO\ is made of a \textit{geology branch} that encodes the geology with factorized Fourier ({F-Fourier}) layers, and a \textit{source branch} that transforms the vector of source parameters $(\bm{x}_s, \bm{\theta}_s)$ into a 4D variable $v_S$ matching the dimensions of the \textit{geology branch} output $v_K$. The outputs of each branch are concatenated after elementary mathematical operations and the remaining factorized Fourier layers are applied. Uplift $P$ and projection $Q_E$, $Q_N$, $Q_Z$ blocks are the same as in the F-FNO: they embed the geology field and project the latent space into the solution space, respectively. Reprinted from~\cite{Lehmann_Gatti_Clouteau_2025}.}
    \label{fig:MIFNO}
\end{figure}

The \MIFNO\ is a neural surrogate model $\mathbb{G}_{\bm{\phi}} : (\bm{a}, \bm{x_s}, \bm{\theta_s}) \mapsto \bm{v}\vert_{\partial \Omega_{top}}$ of the Green's operator, parametrized by weights $\bm{\phi}$, predicting the velocity field $\bm{v}$ at the Earth's surface $\partial \Omega_{top}$, from the sole knowledge of the geology $\bm{a}$ and the source characteristics $\bm{x_s}, \bm{\theta_s}$ \footnote{\url{https://github.com/lehmannfa/MIFNO}}. As shown in Figure~\ref{fig:MIFNO}, the \MIFNO\ is fed with complex heterogeneous three-dimensional (3D) geological fields alongside data detailing earthquake point-source characteristics (position and moment tensor).
The \MIFNO\ consists in a deep neural network of trainable weights $\bm{\phi}$, with two branches encoding the geology field $\bm{a}(\bm{x})$ and source characteristics $\bm{x}_s$, $\bm{\theta}_s$ respectively. The \MIFNO\ backbone architecture is factorized Fourier (F-Fourier) layers~\citep[Factorized Fourier Neural Operator, F-FNO, proposed by][]{Tran_Mathews_Xie_Ong_2023}. Further details are provided in Figure~\ref{fig:MIFNO}. The \MIFNO\ builds on the factorized Fourier Neural Operator (F-FNO, \cite{Tran_Mathews_Xie_Ong_2023}) to learn the velocity field obtained with physics-based numerical simulations. The reference simulations were computed with the high-performance computing open-source software \SEM~\citep{SEM3D_2017}, developed by the CEA, CentraleSup\'elec, CNRS and Institut de Physique du Globe de Paris\footnote{\url{https://github.com/sem3d/sem}}. \SEM\ demonstrated its ability to numerically reproduce high-fidelity fault-to-structure scenarios at regional scale (see, for instance \cite{Touhami_et_al_2022,Castro-Cruz_Gatti_Lopez-Caballero_Hollender_El-Haber_Causse_2022,Fares_CastroCruz_Foerster_LopezCaballero_Gatti_2022}).

\begin{figure}[ht]
    \centering
    \includegraphics[width=\textwidth]{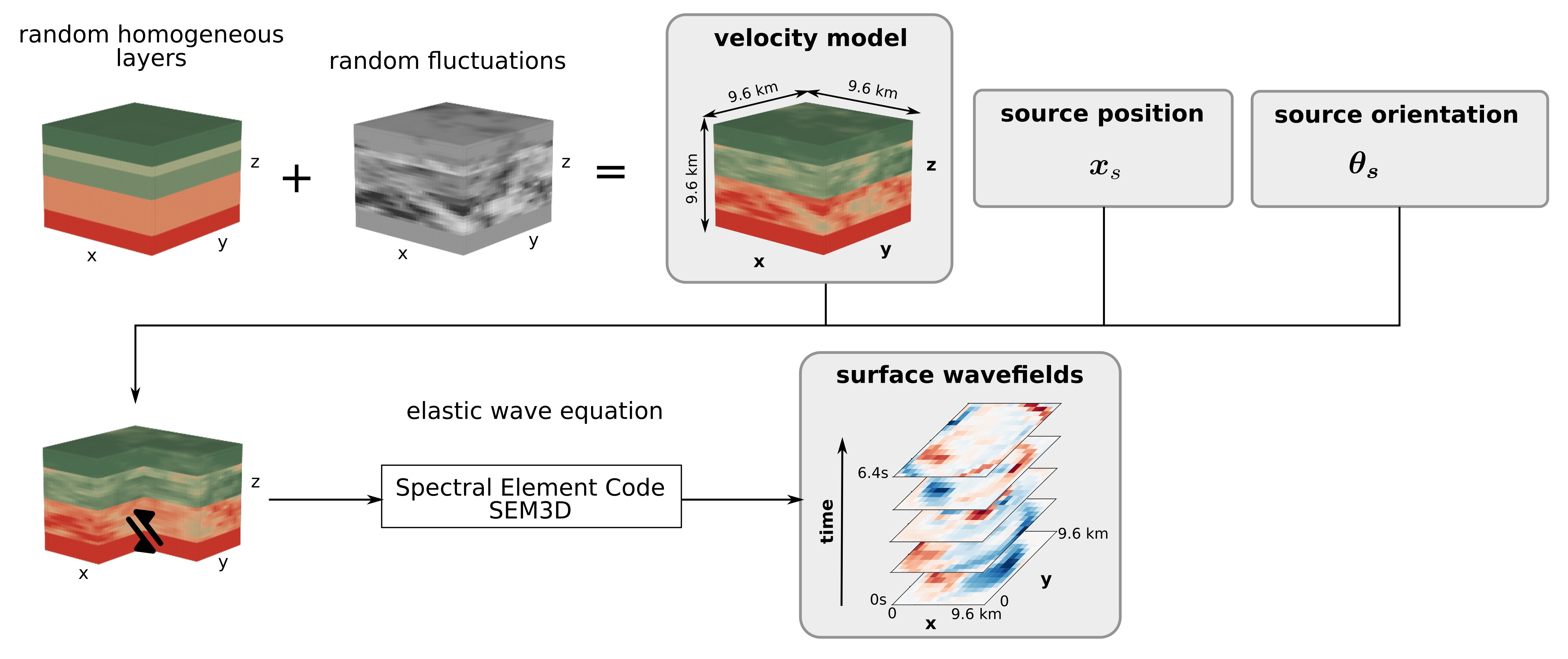}
    \caption{Composition of the \HEMEWS\ database. Velocity models were built from the addition of randomly chosen horizontal layers and heterogeneity drawn from random fields. Combined with the source position and the source orientation, they form the inputs of the neural operator. Outputs of the Spectral Element Code SEM3D are velocity wavefields synthesized at the surface of the domain by a grid of virtual sensors. Reprinted from~\cite{lehmann_synthetic_2024}.}
    \label{fig:HEMEW3D_source_workflow}
\end{figure}

The \MIFNO\ is trained with the open \HEMEWS\ database \citep{lehmann_synthetic_2024} to minimize the error between the predicted surface velocity fields $\bm{\hat{v}}$ and the simulated ones $\bm{v}$. \HEMEWS\ encompasses 30,000 \SEM\ earthquake simulations, conducted in various heterogeneous geologies $\bm{a}(\bm{x})$, incorporating random source positions $\bm{x_s}$ and orientations $\bm{\theta_s}$ (see Figure \ref{fig:HEMEW3D_source_workflow}). The \HEMEWS\ database is also the training database of the \DDPM\ presented in this work.

\cite{Lehmann_Gatti_Clouteau_2025} proved that the \MIFNO\ yields accelerograms that score as good to excellent according to \cite{Kristekova_Kristek_Moczo_2009} Goodness-Of-Fit (GOF) metrics, compared to \SEM\ reference signals. The accuracy of wave arrival times and the propagation of wave fronts is particularly notable, with 80\% of the predictions achieving an outstanding phase GOF score greater than 8. Moreover, the amplitude fluctuations are deemed satisfactory for 87\% of the predictions (with an envelope GOF exceeding 6) and are classified as excellent for 28\% of cases.
MIFNO demonstrates a commendable ability to generalize its predictions for sources located outside the training domain, and it exhibits solid generalization performance when applied to a real-world complex geological scenario characterized by overthrust formations. Additionally, when concentrating on specific regions of interest, transfer learning proves to be an effective strategy to enhance prediction accuracy without incurring significant costs.
However, the prediction accuracy worsens when frequency increases. This is caused by the complex physical phenomena tied to high-frequency features and also reflects the well-known bias in neural networks, which states that small-scale features are more difficult to capture. This limitation can hamper the use of \MIFNO\ in earthquake nuclear engineering applications.

\subsection{Denoising Diffusion Probabilistic Model (\DDPM)}
\label{subsec-ddpm}
\begin{figure}
    \centering
    \includegraphics[width=0.9\textwidth]{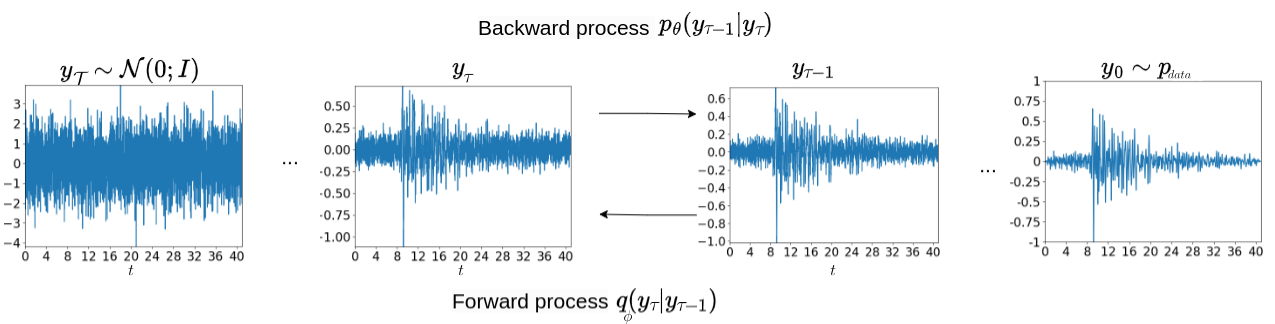}
    \caption{Diffusion process diagram. The forward process is associated to a transition kernel $q_{\phi}(\bm{v}_{\tau}\vert\bm{v}_{\tau-1})$, whereas the backward diffusion process is associated to the transition kernel $p_{\theta}(\bm{v}_{\tau}\vert\bm{v}_{\tau+1})$.}
    \label{fig:Diffusion_photo}
\end{figure}

Our \DDPM\ is based on the original formulation by~\cite{Ho_et_al_2020}. A \DDPM\ exploits Markov Chain Monte Carlo sampling to generate realistic data (in our case, the 3-component seismograms $\bm{v}_0(t)\sim p_{data}$), starting from a sample drawn from the normal distribution $\bm{v}_{\mathcal{T}}(t)\sim\mathcal{N}(\mathbf{0}, \mathbb{I})$. The Markov chain of a diffusion model consists of two discrete processes: the \textit{forward process}, blurring the original data $\bm{v}_0(t)$ till reaching white noise, and the \textit{backward process}, denoising the blurred data $\bm{v}_{\mathcal{T}}(t)$ to a new sample $\bm{\hat{v}}(t)$ that resembles the original data (see Figure \ref{fig:Diffusion_photo}).

$\mathcal{T}$ is the duration of the diffusion process (typically, $\mathcal{T} = 1000$ steps). The forward process consists of generating samples $(\bm{v}_{\tau})_{\tau=1}^{\mathcal{T}}$ with a Gaussian transition probability $q_{\phi}$ that reads:
\begin{equation}
    q_{\phi}(\bm{v}_{\tau}\vert \bm{v}_{\tau-1}) := \mathcal{N}\left( \bm{v}_{\tau} ; \bm{v}_{\tau-1} \sqrt{1 - \beta_\tau}, \beta_{\tau} \mathbb{I} \right),\quad
    q(\bm{v}_{\tau}|\bm{v}_0) := \mathcal{N}\left( \bm{v}_{\tau} ; \sqrt{\bar{\alpha}_{\tau}} \bm{v}_0, (1 - \bar{\alpha}_{\tau}) \mathbb{I} \right)
    \label{eq:forward_process}
\end{equation}
where $ \alpha_{\tau} = 1 - \beta_{\tau} $ and $ \bar{\alpha}_{\tau} = \prod_{s=1}^{\tau} \alpha_s$. The sequence $(\bm{v}_{\tau})_{\tau=1}^{\mathcal{T}}$ is generated by progressively adding Gaussian noise to an initial time series $\bm{v}_0$ selected from the available dataset. A noise variance $ (\beta_{\tau})_{\tau=1}^{\mathcal{T}}$ is predefined by linearly increasing $ \beta_{\tau}$ from 0.0001 to 0.02~\citep{Ho_et_al_2020}. The backward process is defined as the inverse of the forward process and allows for the approximation of $\bm{v}_0$ starting from $ \bm{v}_{\mathcal{T}} \sim p_{\theta}(\bm{v}_{\mathcal{T}}) = \mathcal{N}(\bm{y}_{\tau}; \bm{0}, \mathbb{I})$. The reverse process estimates $ \bm{v}_0 $ by maximizing the likelihood $ p_{\theta}(\bm{v}_0\vert\bm{v}_{\mathcal{T}}) $, defined as the product of transition probabilities:
\begin{equation}
    p_\theta(\bm{v}_{\tau-1}|\bm{v}_{\tau}) := \mathcal{N}\left( \bm{v}_{t-1}; \mu_\theta(\bm{v}_{\tau}, \tau), \sigma_{\tau}^2 \mathbb{I} \right), \quad
    p_\theta(\bm{v}_0|\bm{v}_{\mathcal{T}}) := p(\bm{v}_{\mathcal{T}}) \prod_{\tau=1}^{\mathcal{T}} p_\theta(\bm{v}_{\tau-1}|\bm{v}_{\tau})
    \label{eq:reverse_process}
\end{equation}
where $ \mu_\theta(\bm{v}_{\tau}, \tau) $ and $ \sigma_{\tau}^2 $ are the mean and variance estimated by the model for $\bm{v}_{\tau-1}$.

In practice, a neural network $ \bm{\varepsilon_{\theta}} $ is used to parametrize $ p_\theta(\bm{v}_{\tau-1}|\bm{v}_{\tau})$. This network infers the Gaussian noise $\bm{\varepsilon_{\theta}}(\bm{v}_{\tau}, \tau) $ required to obtain $ \bm{v}_{\tau-1}$ from $\bm{y}_{\tau}$, according to the following expression~\citep{Ho_et_al_2020}:
\begin{equation}
    \bm{v}_{t-1} = \frac{1}{\sqrt{\alpha_{\tau}}} \left( \bm{v}_{\tau} - \frac{1 - \alpha_{\tau}}{\sqrt{1 - \bar{\alpha}_{\tau}}} \bm{\varepsilon_{\theta}}(\bm{v}_{\tau}, t) \right) + \sqrt{\beta_{\tau}} \bm{z}_{\tau}, \quad \bm{z}_{\tau} \sim \mathcal{N}(\bm{0}, \mathbb{I}).
    \label{eq:diffusion_process}
\end{equation}
The term $\bm{\varepsilon_{\theta}}(\bm{v}_{\tau}, \tau)$ represents the estimation of the noise level contained in $\bm{v}_{\tau}$ at time step $\tau$. The challenge of diffusion models is thus to estimate the noise generated by the forward process, thereby enabling the recovery of $\bm{v}_0$ from $\bm{v}_{\mathcal{T}}$. The \DDPM\ neural network $\bm{\varepsilon_{\theta}}$ ($\bm{\theta}$ represent the neural network trainable weights) is trained to minimize the following loss:
\begin{equation}
    \mathit{L}(\bm{\theta})=\mathbb{E}_{\tau\sim \mathcal{U}([1,\mathcal{T}]),\bm{v}_0\sim p_{data},\bm{\varepsilon}_{\tau}\sim\mathcal{N}(\bm{0},\mathbb{I})}\left[\Vert\bm{\varepsilon}_{\tau} - \bm{\varepsilon_{\theta}}\left(\sqrt{\bar{\alpha_{\tau}}}\bm{v}_0 + \sqrt{1- \bar{\alpha_{\tau}}}\bm{\varepsilon};\tau\right) \Vert^{2}\right]
    \label{eq-eq5}
\end{equation}
In other words, $\bm{\varepsilon_{\theta}}$ plays the role of a denoiser that progressively transforms the noisy samples $\bm{v}_{\tau}$ into cleaner samples $\bm{v}_{\tau-1}$, by matching the estimated white noise added by the forward process.
\cite{Gabrielidis_Gatti_Vialle_Jacquet_2024} adopted a UNet as the $ \bm{\varepsilon_{\theta}} $ denoiser of 3-component accelerograms. This \DDPM\ was trained to generate broadband (0-30Hz) signals, while being conditioned on the low-pass filtered signals (0-1Hz). In this manner, $\bm{\varepsilon_{\theta}}$ learns to conditionally denoise a white noise sample, being forced to match the conditioning input at low-frequency. %The latter is fed to $\bm{\varepsilon_{\theta}}$ by inserting layers adaptation of the {ResBlockDown} and {ResBlockUp} blocks proposed by~\cite{Brock_Donahue_Simonyan_2019}, as well as cross-attention mechanisms~\citep{Vaswani_Shazeer_Parmar_Uszkoreit_Jones_Gomez_Kaiser_Polosukhin_2023}. At inference time, such condition guiding the diffusion process can be obtained by other means, such as earthquake numerical simulation or surrogate model synthetics (as the \MIFNO's output), provided that such low-frequency seismic signal corresponds to realistic earthquake ground motion spectrum.

\subsection{Enhancing \MIFNO\ with \DDPM}
\begin{figure}[ht!]
    \centering
    \includegraphics[width=\textwidth]{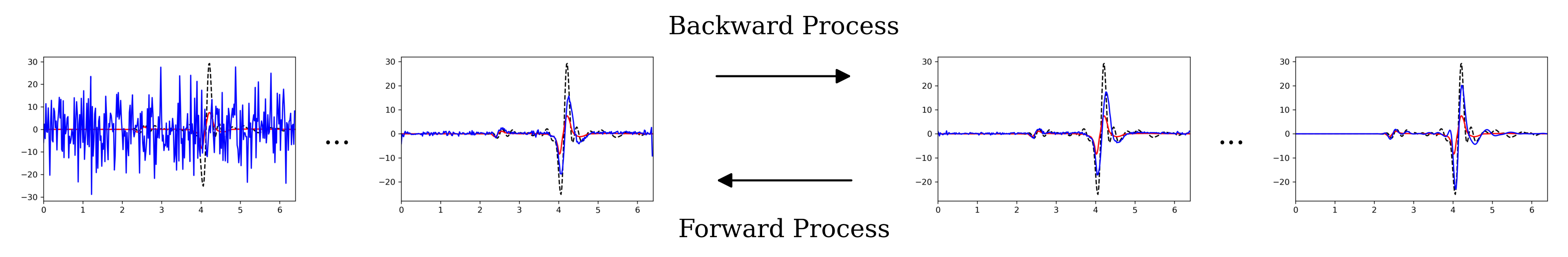},
    \caption{Forward and backward diffusion process conditioned by the \MIFNO\ output (red). The \SEM\ numerical solution, representing the target, is shown in black, while the \DDPM\ output is shown in blue.}
    \label{fig:diffusion_fw_bw_mifno}
\end{figure}

Our study takes inspiration from~\cite{Oommen_Bora_Zhang_Karniadakis_2024}, who combine diffusion models and neural operators to enhance the resolution of turbulent structures. In the present work, the \MIFNO\ predicts the ground motion velocity signals that serve as the conditioning of the \DDPM. Since the low-frequency parts of the \MIFNO\ predictions were shown to be close to the reference simulations~\citep{Lehmann_Gatti_Clouteau_2025}, we argue that they contain relevant physical information to guide the diffusion process. It is important to note that the \MIFNO\ is frozen in this work (\textit{i.e.}, its weights are not updated). The \DDPM\ is trained to generate velocity time series from the \HEMEWS\ database, conditioned on the corresponding velocity prediction by the \MIFNO\ (Fig. \ref{fig:diffusion_fw_bw_mifno}).

Our \MIFNODDPM\ is trained on 19,200 realizations from the \HEMEWS\ database. From each realization, 32 sensors are randomly selected to obtain the 3-component velocity time series. This leads to 614,400 3-component velocity time series of duration 6.4s to train the \MIFNODDPM. Training is done for 100 epochs, with a batch size 256, which took 13.29h on four A100 GPUs. A cosine annealing learning rate scheduler was used with 10 restarts, each cycle starting with learning rate 3e-4 and ending with 0.

At inference time, one entire wavefield prediction takes around 50ms and it can be enhanced with \DDPM\ on 512 sensors at once in 3 minutes on one A100 GPU.

\section{Results}
\label{sec-results}
\begin{figure}[ht!]
    \centering
    \subfloat[]{\includegraphics[width=0.32\textwidth]{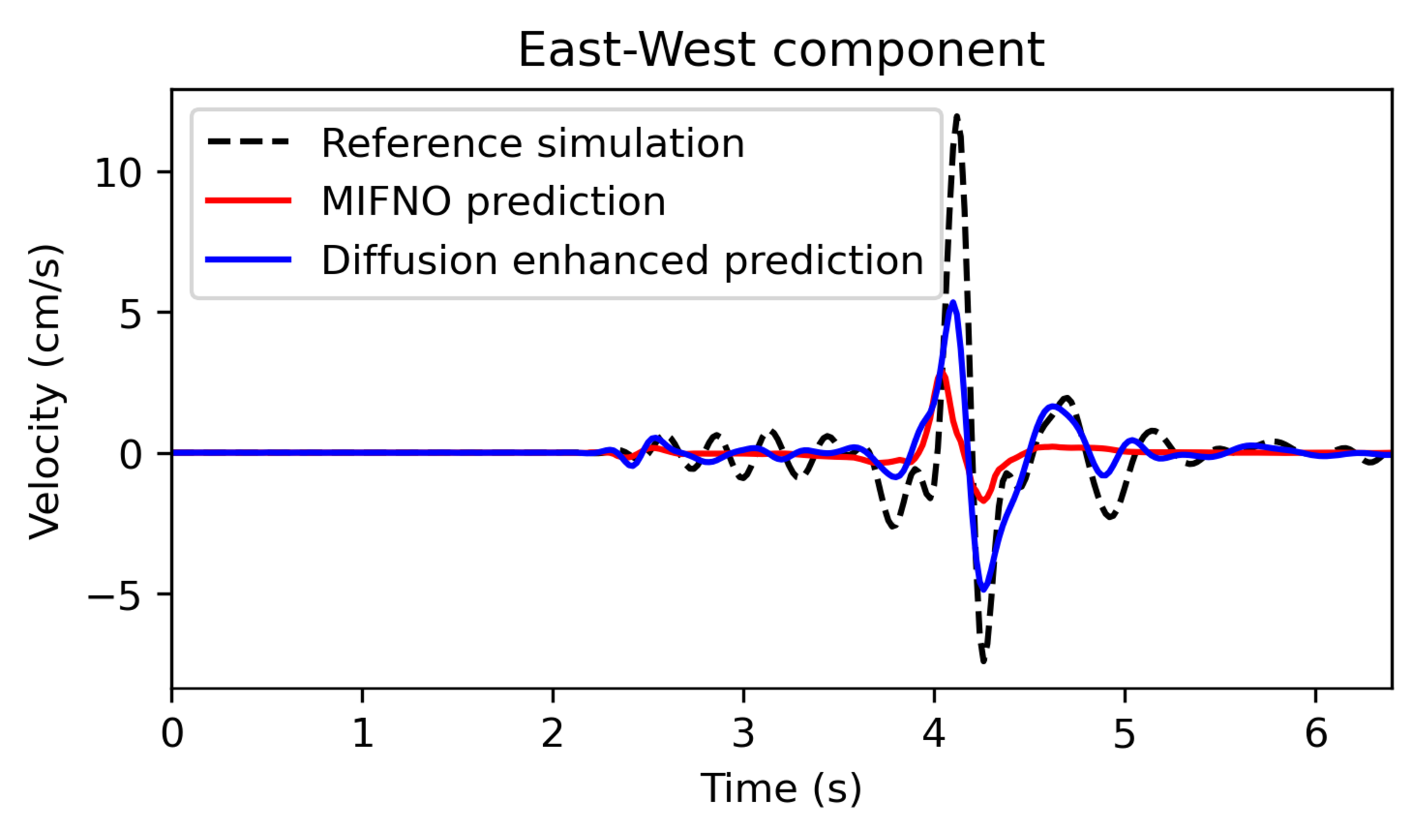}}
    \subfloat[]{\includegraphics[width=0.32\textwidth]{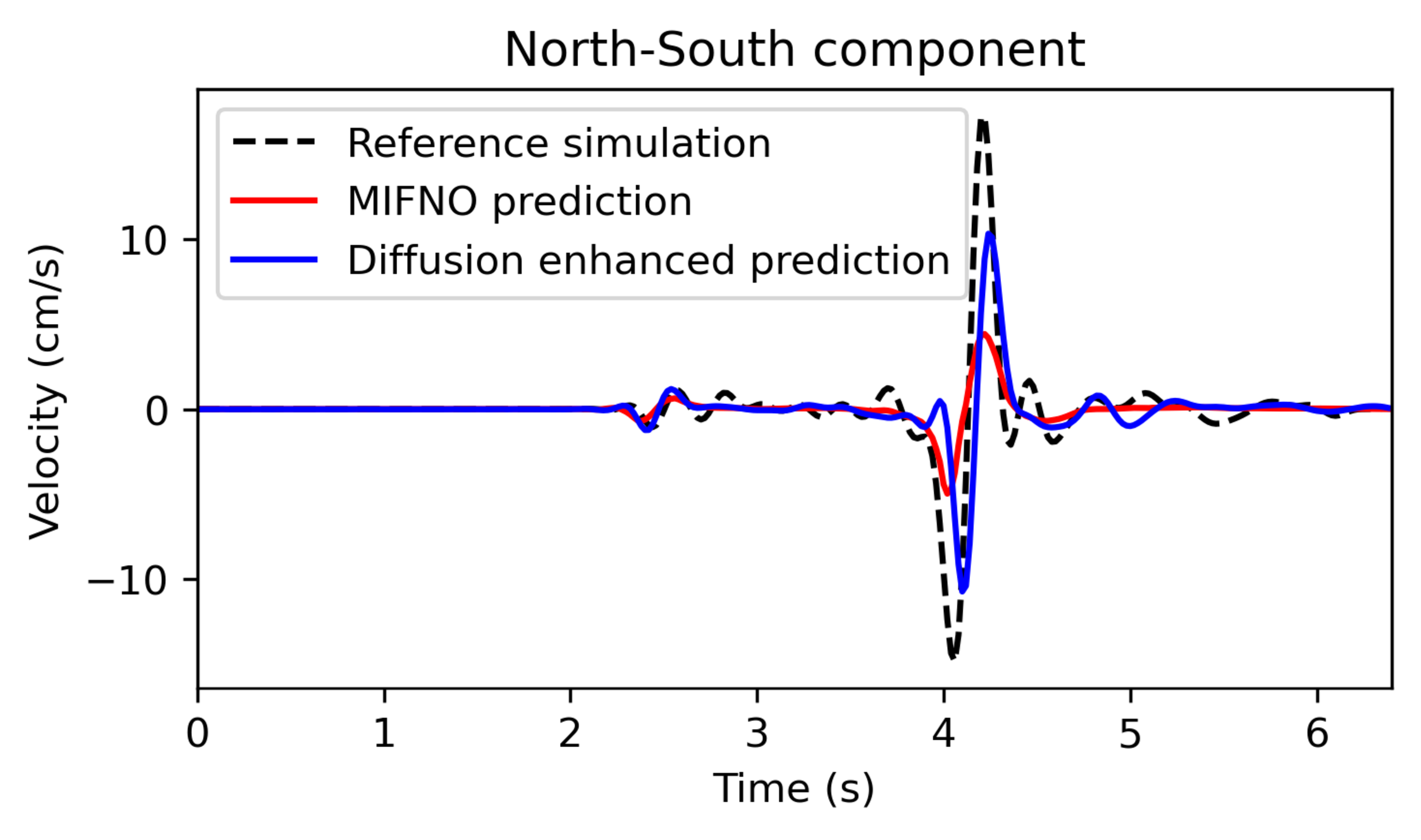}}
    \subfloat[]{\includegraphics[width=0.32\textwidth]{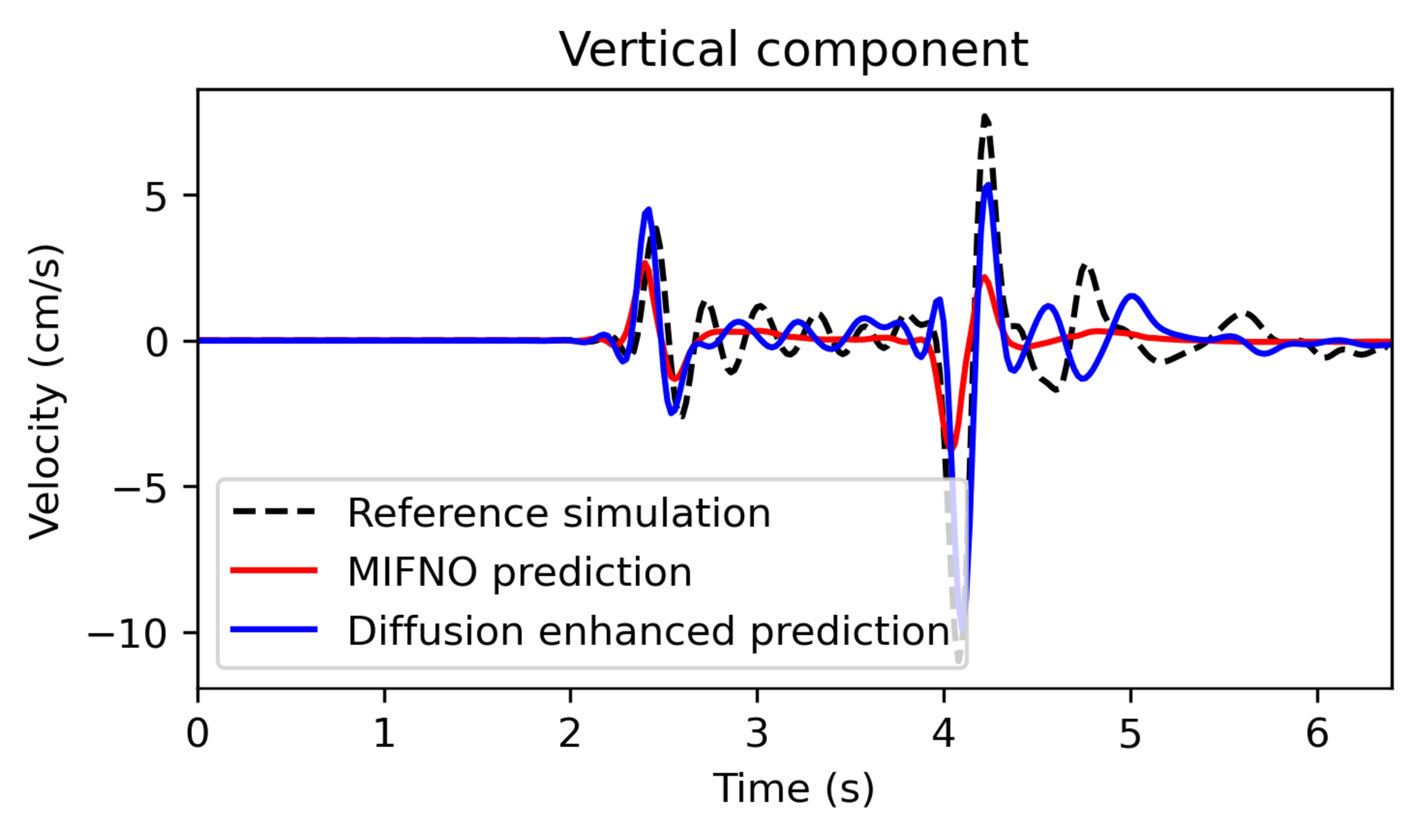}}\\
    \subfloat[]{\includegraphics[width=0.32\textwidth]{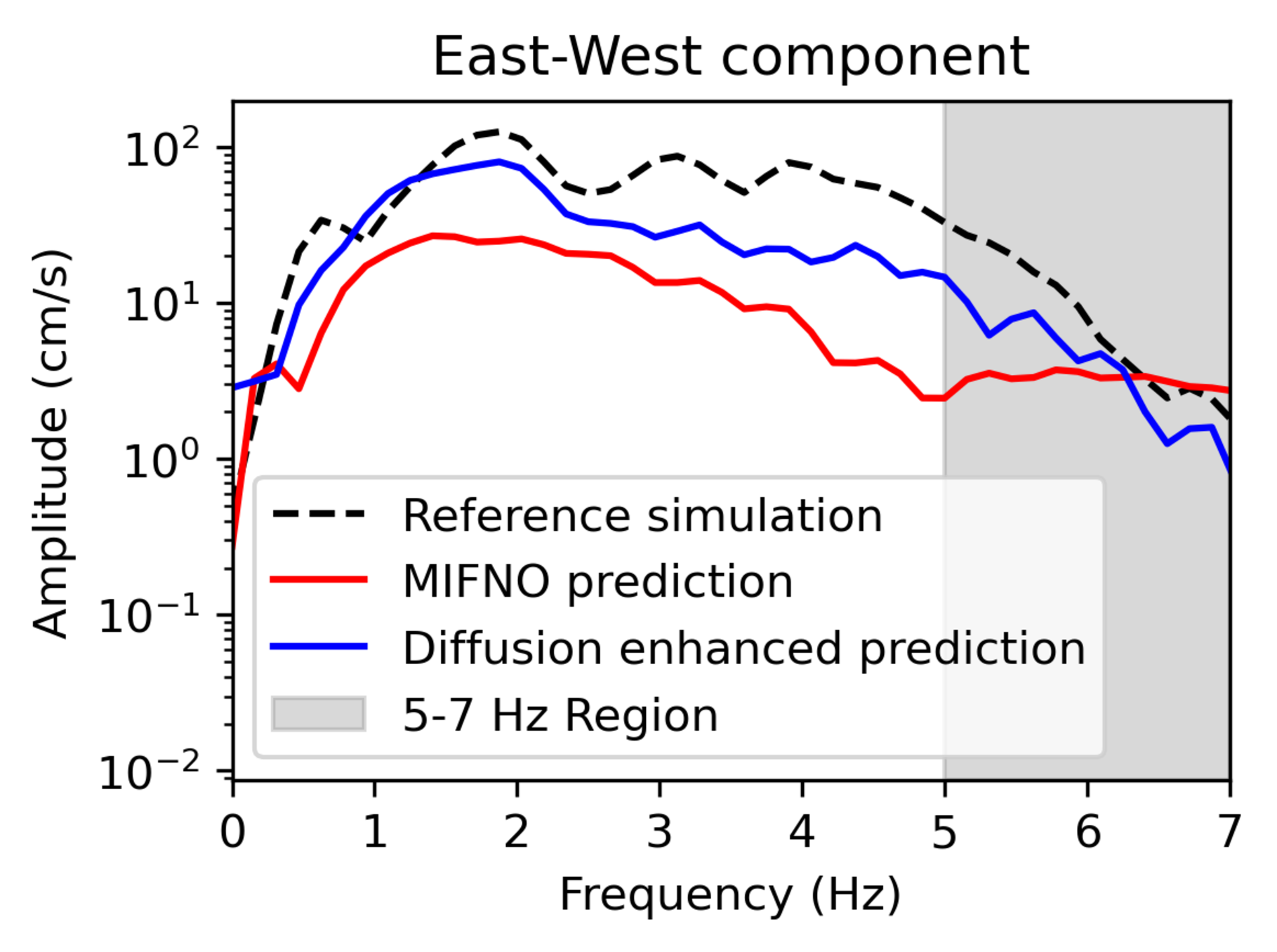}}
    \subfloat[]{\includegraphics[width=0.32\textwidth]{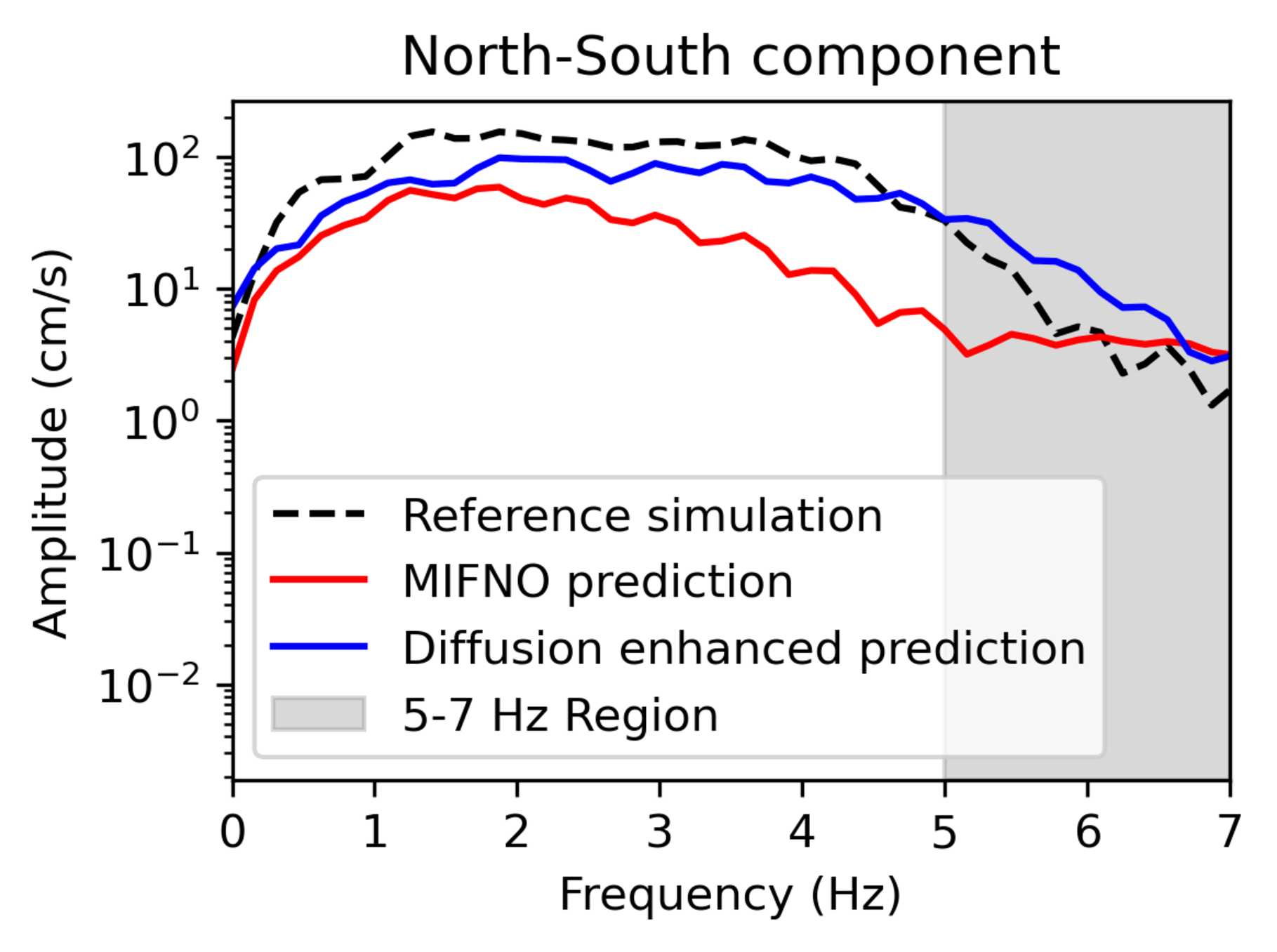}}
    \subfloat[]{\includegraphics[width=0.32\textwidth]{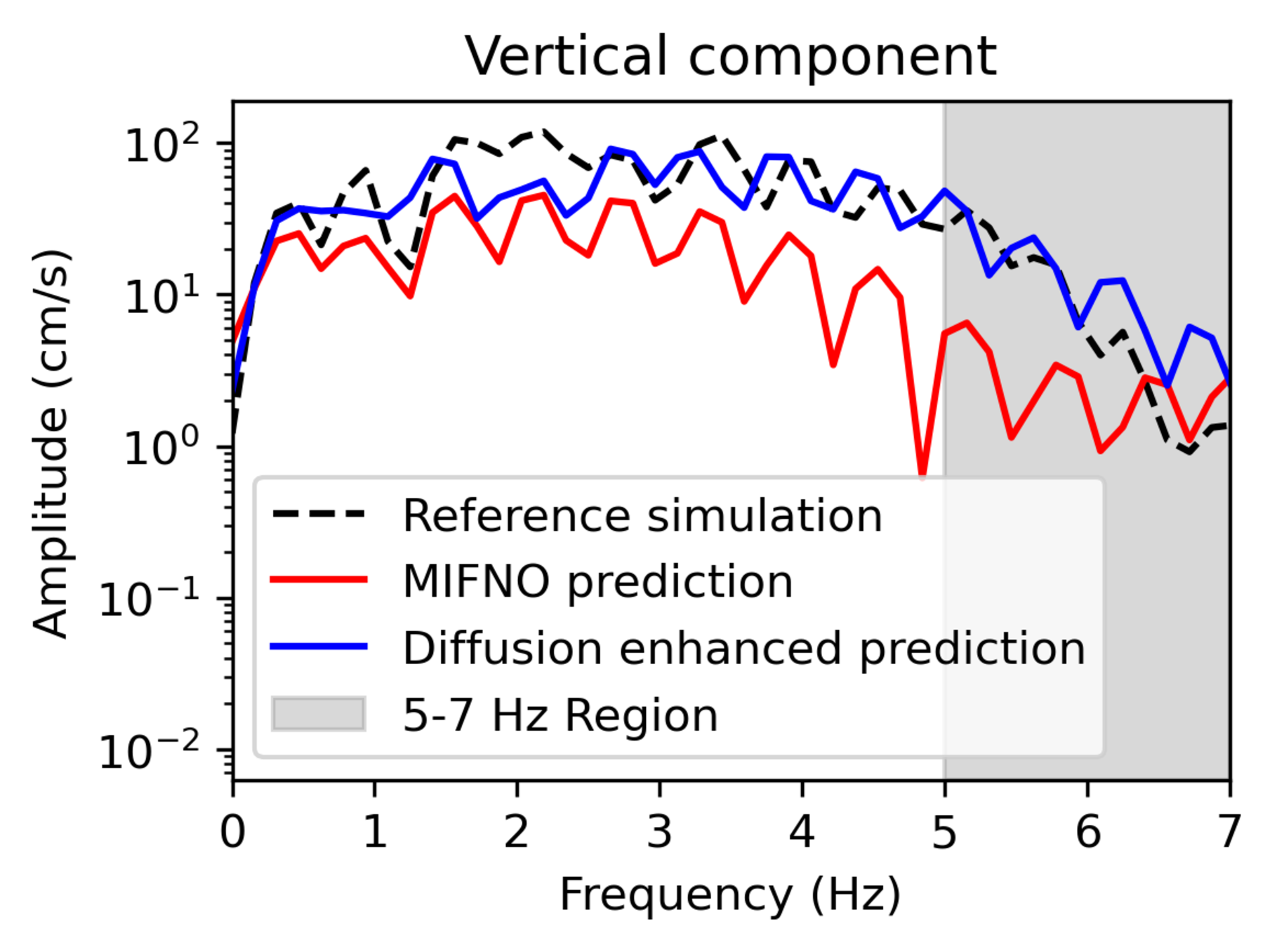}}
    \caption{Comparison between the reference numerical simulation (black dashed line), the \MIFNO\ prediction acting as conditioning (red line), and the signal enhanced by \DDPM\ (blue line). The top panels (a, b, c) show the 3-component synthetic seismograms, while the bottom panels (d, e, f) show the corresponding spectra.}
    \label{fig:mifno_diff}
\end{figure}

Figure \ref{fig:mifno_diff} (a-c) illustrates that the velocity time series obtained with \MIFNO\ + \DDPM\ (blue line) is closer to the reference simulation (black dashed line) than the original \MIFNO\ prediction (red line). This shows that enhancing the surrogate model with the diffusion process improves the accuracy of the predictions. It is important to observe that the enhanced time series preserve the wave arrivals that were correctly predicted by the \MIFNO. This indicates that the \MIFNO\ conditioning is effective in transmitting the physical information of the signal and that the \DDPM\ preserve this information.

When looking at the spectral representations in Fig. \ref{fig:mifno_diff} (d-f), one can observe that the \DDPM\ improves the agreement between the predictions and the reference simulations in the entire frequency range, thereby mitigating the spectral bias of \MIFNO.

\begin{table}[h]
    \caption{Accuracy metrics of the original \MIFNO\ (first line) and \MIFNO\ enhanced by \DDPM\ (second line). 3000 test samples are compared with the reference simulations. The best metrics are indicated in bold. rMAE = relative Mean Absolute Error, rRMSE = relative Root Mean Square Error, rFFT = bias in Fourier coefficients between 0 and 1 Hz (low), 1 and 2 Hz (mid), 2 and 5 Hz (high), EG = Envelope GOF, PG = Phase GOF. More details on the metrics can be found in \cite{lehmann3DElasticWave2024}.}
    \centering
    \begin{tabular}{|c|c|c|}
        \hline
        Metric                   & \MIFNO\                    & \MIFNO\ + \DDPM\             \\
        \hline
        rMAE                     & \textbf{0.16 $\pm$ 0.06} & 0.20 $\pm$ 0.09          \\
        rRMSE                    & \textbf{0.30 $\pm$ 0.11} & 0.34 $\pm$ 0.18          \\
        rFFT\textsubscript{low}  & -0.27 $\pm$ 0.19         & \textbf{0.04 $\pm$ 0.32} \\
        rFFT\textsubscript{mid}  & -0.36 $\pm$ 0.23         & \textbf{0.08 $\pm$ 0.45} \\
        rFFT\textsubscript{high} & -0.44 $\pm$ 0.27         & \textbf{0.11 $\pm$ 0.60} \\
        EG                       & 6.15 $\pm$ 1.00          & \textbf{6.16 $\pm$ 1.12} \\
        PG                       & 6.65 $\pm$ 1.07          & \textbf{7.52 $\pm$ 1.08} \\
        \hline
    \end{tabular}
    \label{tab:metrics}
\end{table}

\begin{figure}[ht!]
    \centering
    \subfloat[]{\includegraphics[width=0.5\textwidth]{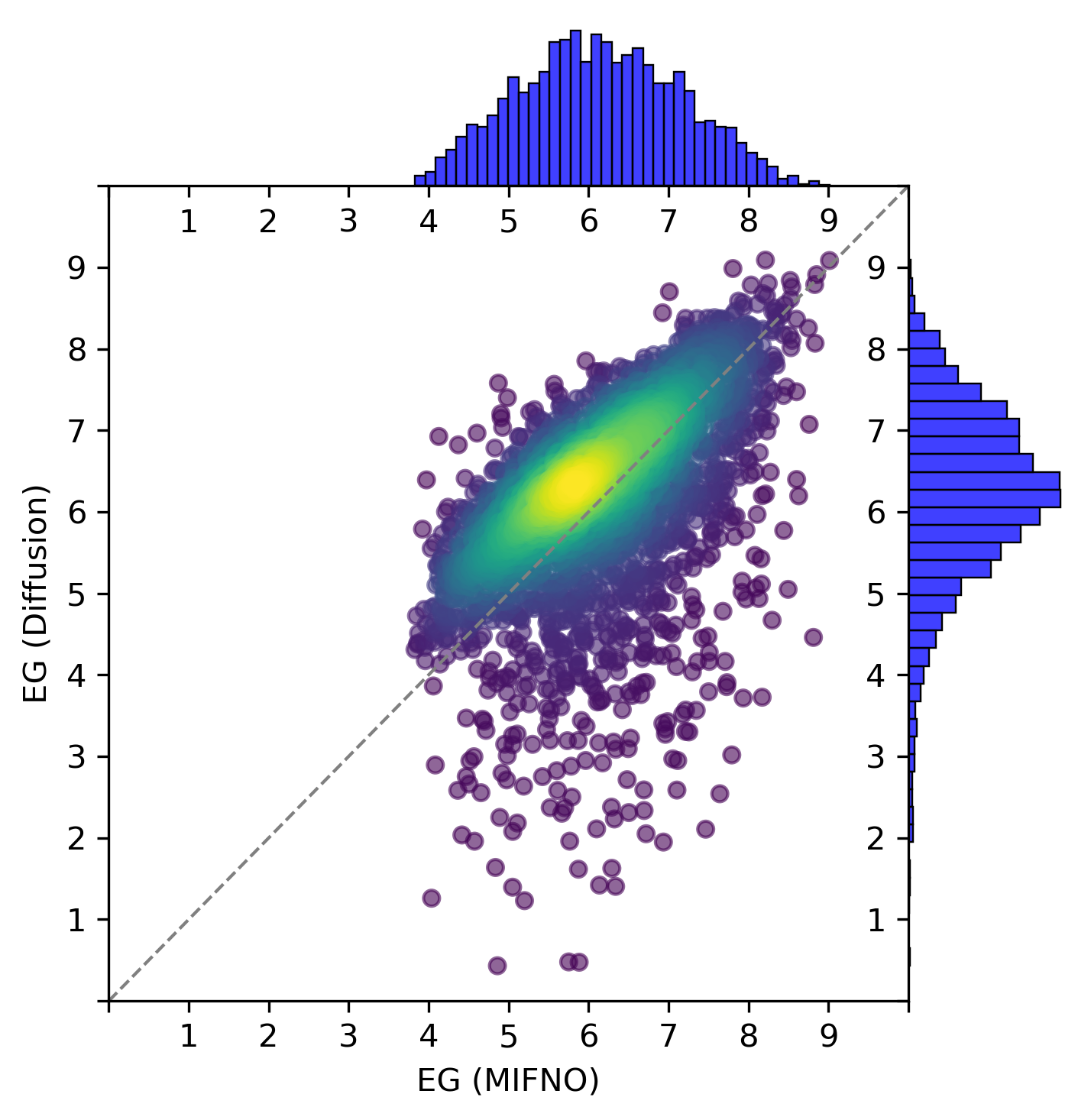}\label{sfig:EG_mifno_vs_diffusion}}
    \subfloat[]{\includegraphics[width=0.5\textwidth]{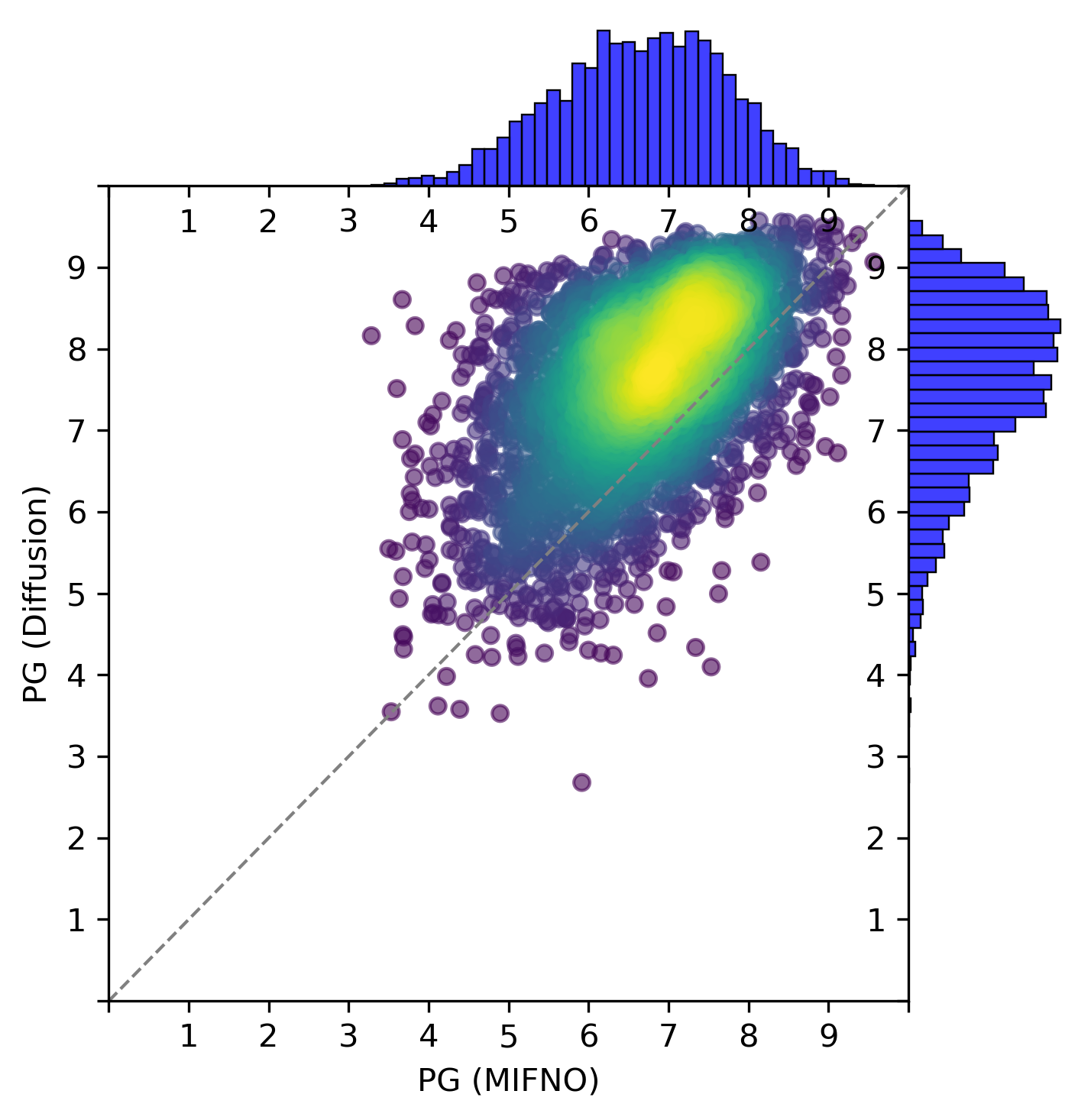}\label{sfig:PG_mifno_vs_diffusion}}
    \caption{GOF scores (from 0 to 10, 10 being a perfect agreement) comparing the reference numerical simulation with the \MIFNO\ (horizontal axis) and \MIFNO~+~DDPM (vertical axis). (a) Envelope GOF; (b) Phase GOF. All time series were low-pass filtered in the 0-5 Hz frequency range for all the three components. The GOF score represents the average GOF across the three components. The color in the scatter plot shows the points density.}
    \label{fig:GOF}
\end{figure}

These visual observations are corroborated by the metrics computed on 3000 test samples. The \MIFNO\ + \DDPM\ preserves the physical consistency of the \MIFNO\ since the enveloppe and phase GOF (resp. EG and PG) are better than the \MIFNO. The improvement is especially visible on the phase GOF. These results are further illustrated in Fig. \ref{fig:GOF} where the large majority of points lie above the diagonal.

The \DDPM\ corrects the \MIFNO\ pre-conditioning across the entire 0-5 Hz frequency range as it reduces the frequency biases (in absolute value) given in Tab. \ref{tab:metrics}. One can however notice that the relative MAE and RMSE are slightly higher for the \MIFNO\ + \DDPM\ than for the \MIFNO. This can be related to some residual noise that affects these point-wise metrics while the physically-based metrics (\textit{i.e.}, frequency biases and GOF) are not sensitive to these small errors.

Figure \ref{fig:GOF} also indicates some samples where the diffusion model degrades the envelope GOF. This happens in most cases when the \MIFNO\ amplitude is very small and the \DDPM\ is not able to extract enough physical information from the \MIFNO\ conditioning.

\section{Discussion and Conclusions}
\label{sec-discussion_and_conclusions}
In this paper, we foreshadowed the appealing perspective of combining neural operators and diffusion models in computational earthquake engineering. This strategy addresses an open challenge in this domain: the need for a fast earthquake simulator which is also accurate enough in the high-frequency range to be employed to design critical infrastructures. We combined the fast inference of the physics-based \MIFNO\ surrogate model with the high-frequency accuracy of the \DDPM\ generative model. This combination improves the accuracy of the synthetic seismograms while preserving the physical features learned by the \MIFNO.

However, as per now, it must be noted that the seismograms generated by our \MIFNO~+~DDPM are not yet accurate enough for practical applications. One possible explanation is that the \DDPM\ treats the seismograms from the same realization (\textit{i.e.}, same source and same geology but observed at two different sensors) independently. Therefore, the spatial correlation between them is neglected. Explicitly introducing such correlation, acknowledging the difference among within- and between-event realizations, would steer the \DDPM\ towards learning complex spatio-temporal patterns. Further work is required to extend the current \DDPM\ architecture.

Nevertheless, from a practical perspective, the advantage of our proposed strategy lies in the possibility of quickly inferring approximate earthquake ground motion regional response, for any geology and any source, without running expensive numerical simulations. Moreover, the residual gap in accuracy presumably lies within the epistemic uncertainty margins of 0-5 Hz numerical simulations compared to recorded seismograms. With this tool at hand, we make progress towards unlocking physics-based probabilistic seismic hazard assessment. We envision the possibility of relegating cumbersome yet accurate earthquake numerical simulations to offline database enrichment, adopted to progressively fine-tune our \MIFNODDPM\ surrogate model. The repercussions on the design procedures of nuclear facilities are potentially disruptive.

\section*{Acknowledgements}
This work was performed using HPC/AI resources from GENCI-IDRIS (Grant A0170414346).
% An abstract may include references. In-text references should consist of last names of authors and the year, for example, \cite{hughes1969}, \cite{cook2002}, \cite{chang1987}, \cite{ubc1988}, and \cite{frater1992}.
% \bibliographystyle{unsrt}
\bibliographystyle{plainnat}
\bibliography{references}
% \printbibliography

\end{document}